\documentclass{article}
\usepackage{spconf,amsmath,graphicx}

\usepackage{color}
\usepackage{array}

\usepackage{float}
\usepackage{url}
\usepackage{enumerate}
\usepackage{amsfonts}
\usepackage{enumitem}
\usepackage{subcaption}
\newcommand{\etal}{\mbox{\emph{et al.\ }}}

\newcommand{\eg}{\mbox{\emph{e.g.\ }}}

\newcommand{\comment}[1]{}
\newcommand{\T}{{\scriptscriptstyle \top}}

\def\RR{\mathbb{R}}

\title{Classification and Detection in Mammograms with Weak Supervision via Dual Branch Deep Neural Net }
\name{Ran Bakalo$^{1,2}$ \qquad Rami Ben-Ari$^{2}$ \qquad Jacob Goldberger$^{3}$}

 \address{$^{1}$ Department of Computer Science, University of Haifa, Israel, 
     $^{2}$ IBM Research, Haifa, Israel \\
     $^{3}$ Faculty of Engineering, Bar Ilan University, Israel}
\usepackage{tikz}
\usepackage{textcomp}
\usepackage{hyperref}
\usepackage{lipsum}

\newcommand{\copyrighttext}{%
  \footnotesize \textcopyright 2019 IEEE. Personal use of this material is permitted.
  Permission from IEEE must be obtained for all other uses, in any current or future
  media, including reprinting/republishing this material for advertising or promotional
  purposes, creating new collective works, for resale or redistribution to servers or
  lists, or reuse of any copyrighted component of this work in other works.}
\newcommand{\acopyrightnotice}{%
\begin{tikzpicture}[remember picture,overlay]
\node[anchor=south,yshift=10pt] at (current page.south) {\fbox{\parbox{\dimexpr\textwidth-\fboxsep-\fboxrule\relax}{\copyrighttext}}};
\end{tikzpicture}%
}

\begin{document}
%
\maketitle
\acopyrightnotice
\begin{abstract}
The high cost of generating expert annotations, poses a strong limitation for supervised machine learning methods in medical imaging. Weakly supervised methods may provide a solution to this tangle.
In this study, we propose a novel deep learning architecture for multi-class classification of mammograms according to the severity of their containing anomalies, having only a global tag over the image. The suggested scheme further allows localization of the different types of findings in full resolution. The new scheme contains a dual branch network that combines region-level classification with region ranking. We evaluate our method on a large multi-center mammography dataset including $\sim$3,000 mammograms with various anomalies and demonstrate the advantages of the proposed method over a previous weakly-supervised strategy.
\end{abstract}

\section{Introduction}
\label{sec:intro}
Screening mammograms (MG) are the first line of imaging for the early detection of breast cancer, that raise the survival rate, but place a massive workload on radiologists. Although MG provides a high resolution image, its analysis remains challenging, due to tissue overlaps, high variability between individual breast patterns, and subtle malignant findings.

A naive approach for classifying mammograms suggests training a detector from local annotations \cite{DBLP:journals/corr/RibliHUPC17, DBLP:journals/corr/LotterSC17}, and then classifying the images according to the most severe detection. However, this requires instance level annotations for every finding in the image which is tedious, costly and impractical for massive datasets. A common technique to combat this problem is using weakly supervised paradigms, where labels are given only at the image-level (\eg knowing a certain image contains a malignant finding but without localization). Some weakly supervised methods can further provide localization, as discriminative regions. This can be used as explanation behind the model decision particularly where the source of discrimination between the classes is a-priori unknown.

Previous weakly supervised approaches for mammogram classification  \cite{choukroun2017mammogram,DBLP:conf/miccai/ZhuLVX17} used the popular multi-instance learning (MIL) concept.  Hwang \etal \cite{hwang2016self} used a CNN with classification and localization branch with a linear combination loss. The shortcomings of these methods include: restriction to binary classification, processing of harshly downsized images \cite{hwang2016self,DBLP:conf/miccai/ZhuLVX17}, implicit detection paradigm via max-pooling operation \cite{choukroun2017mammogram,DBLP:conf/miccai/ZhuLVX17} and localization at low resolution \cite{hwang2016self,DBLP:conf/miccai/ZhuLVX17}. Recent studies suggest that applying explicit {\it data-driven} detection in parallel to classification yields improved performance \cite{bilen2016weakly}. 

In this study we explore a new data-driven strategy for weakly supervised learning with a novel dual-branch deep CNN model. Our architecture is composed of two streams, acting on image regions, one for classification and the other for detection. The classification compares association of regions with certain class \eg normal, benign or malignant. In the detection branch however, the scores of regions are compared to others, at each class. Hence, the classification branch predicts which class to associate to a region, whereas the detection branch selects which regions are more likely to contain a finding. 

Our approach is  inspired  by \cite{bilen2016weakly} (applied on natural images) but differs and generalizes the existing methods by two main factors: 1)  We don't use any unsupervised region proposal in our scheme as it is commonly unavailable in MG. 2) We change the architecture in \cite{bilen2016weakly} and extend the classification stream with additional class but without any  detection counterpart, in order to handle images lacking any object (finding). The extension for handling images without objects is equivalent to normal radiology images without any pathology. The additional normal class changes the score distribution for regions, impacting the global score used for image classification. Similar to \cite{jiang2017optimizing}, we further connect between the branches by adding the information from the classification branch to guide the detection to the most relevant regions. Our model is capable of multi-class classification and detection that provides anomaly localization in full resolution. As such we compare our method with previous approaches in both classification and detection, showing improved performance. 
\section{Mammogram classification algorithm}
\begin{figure*}[!ht]
\centering
\includegraphics[width=15 cm, height=5 cm]{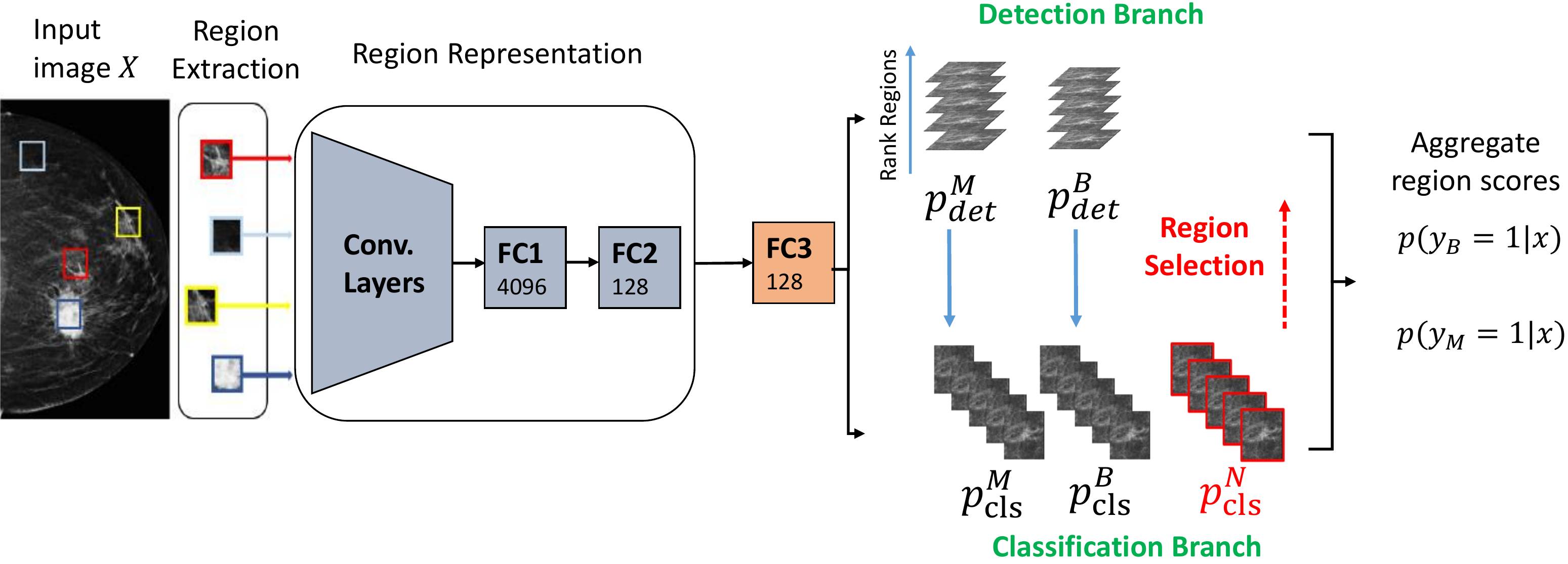}
\caption{Architecture overview.  The novel elements are in red. Our new scheme has an additional class in the classification branch ($p_{cls}^N$ ) with no associated detection. This allows filtering of normal regions to obtain a low global score for a normal breast as desired. The region selection further boosts performance.}
\label{fig:WeakArchitecture}
\end{figure*}

Generally, radiologists divide regions in MG into three classes by severity order:  normal tissues that can be viewed as background, lesions appearing to be benign and finally malignant findings. Our goal is to automatically classify the image into one of these classes. We suggest a novel weakly supervised and deep learning model composed of two branches as described in Fig. \ref{fig:WeakArchitecture}. The branches represent: (1) a {\it classification} that computes probabilities for classes \eg malignant, benign with addition of a normal class for each image region, (2) a {\it detection} branch that compares all regions and ranks them according to the belief of containing a pathology. It is therefore a proxy to detection. The branches are then combined downstream to obtain an image-level decision for the presence of malignant and/or benign findings (see Fig. \ref{fig:WeakArchitecture}).
In the following sections we describe in detail the building blocks of our scheme.

\textbf{Region extraction.}
 An image $x$, is first decomposed into $m$ regions denoted by $x_1,...,x_m$. To this end, we used a sliding window of $224 \times 224$ overlapping regions (with $112 \times 112$ stride) in the breast excluding the axilla (which should be handled separately due to the different nature of the findings in this region). Through several layers of CNN each region is encoded into a feature representation. Due to the relatively small training dataset, we employ a two-stage method. In the first stage, we apply a transfer-learning approach by using a pretrained VGG128 network \cite{Chatfield14}, trained on the ImageNet dataset \cite{deng2009imagenet}. In our model, we extract CNN codes from the last hidden layer such that  $x_i \in \RR^{128}$ is the feature vector representation of the $i$-th region (other VGG configurations and layers yielded inferior results). Then, we process each region separately by a fully connected (FC) layer (FC3 in Fig. \ref{fig:WeakArchitecture}).

\textbf{Classification branch.} Each region is classified (soft decision) as $N$, $B$ or $M$, associated with healthy (normal) tissue, benign or malignant abnormality. To this end we use the following softmax layer (corresponding to lower branch in Fig. \ref{fig:WeakArchitecture}):
\begin{equation}
\begin{split}
 &p_{\mbox{cls}}(c|x_i) =  \frac{ \exp (w_c^{\T} x_i)} { \sum_{d\in\{N,B,M\}}  \exp (w_d^{\T} x_i)}  \\
 &c\in\{N,B,M\}, \quad i=1,\ldots,m
\end{split}
\label{eq:class_br}
\end{equation}
such that $w_N, w_B$ and $w_M$ are the parameters of the respective classifier. Note that the same classification parameters are used across all the regions in the image.

{\bf Region selection.} It makes sense to connect between the decision processes of classification and detection. For example, if a region is strongly classified as $M$ we would like that the malignancy detector will favor this region. We therefore impose a soft alignment between the branches by making the classification branch selecting regions for detection. Note that the classification branch includes a separate normal class, thus allowing to filter out normal regions from the detection branch (by assigning them low probabilities). We formalize this intuition by a region selection scheme. For each abnormality class in the detection branch we consider the corresponding $k$ top scoring regions, removing the regions that obtained low classification scores. As the number of selected regions we chose $k=10$. This implicit region selection scheme guides the detector to focus on the most likely abnormal regions in one hand and forces the network to improve classification in other hand (due to loss amplification on those errors). For normal breast images, where all regions are of class $N$, erroneously classified regions act as {\it hard negative mining}.

\textbf{Detection branch.} In parallel, we compute the relevance of each region as an abnormality by ranking (upper branch in Fig. \ref{fig:WeakArchitecture}) . There is a separate ranking for each class; in this study one for malignant and the other for benign. Since the prevalent normal regions are not considered as abnormality, these regions are treated as ``background" and therefore not associated with a separate detection class. The detection result is therefore a distribution over all the regions in the image:
\begin{equation}
\begin{split}
&p^c_{\mbox{det}}(i|x) = \frac{ h_c(i) \exp(u_c^{\T} x_i)}{\sum_{j=1}^m h_c(j) \exp ( u_c^{\T} x_j) } \\  
&c\in\{B,M\}, \quad i=1,\ldots,m
\end{split}
\label{eq:det_br}
\end{equation}
such that $u_B$ and $u_M$ are the parameter-set of the abnormality detectors and $h_c(i)$ is a $k$-hot vector indicating the $k$-top scoring regions in the associated classification branch. Note that for certain class $c$, Eq. \eqref{eq:det_br} is equivalent to ranking of $i$-th region in image $x$ with respect to other regions.

\textbf{Image level decision.} Given the region-level classification results and the region detection distribution, we can now evaluate the image-level class.
Let $(y_M,y_B)$ be a binary tuple indicator whether an image contains a malignant and/or benign finding, respectively. Note that this type of tuple labeling allows tagging images of class $N$ by $(0,0)$ and those with both $M$ and $B$ findings by $(1,1)$.
The posterior distributions of $y_M$ and $y_B$  are obtained as a weighted average of the local decisions:
\begin{equation}
p(y_c=1|x) = \sum_{i=1}^m p^c_{\mbox{det}} (i|x) p_{\mbox{cls}}(c|x_i),
\quad c\in\{B,M\}.
\label{eq:image_decision}
\end{equation}
Since in many medical applications such as mammography the most prevalent cases are normal without any finding, we extended the method in \cite{bilen2016weakly} by adding a normal ($N$) classifier in the classification branch. This is a novel generalization to previous modeling in \cite{bilen2016weakly} where each object class is associated with a corresponding detector. Normal images in our model are then discriminated by having low probability for all types of abnormalities. The addition of $N$ classifier further reduces false positives in detection (localization) as normal regions gain low scores (instead of uniform scores over all classes of abnormalities). The probability for an image to be normal can be obtained via the joint probability $p(y_M=0,y_B=0|x)$.
 
\textbf{Training.} 
The network output are soft decisions for each train image $x(t)$ as containing any type of abnormality \eg $M$ or $B$, represented by the tuple label $\left(y_M(t),y_B(t)\right)$. We set the following cross-entropy loss function for the network training:
\begin{equation} \label{eq:MC_loss}
L(\theta) =  -\sum_{c \in \{M,B\}} \sum _{t=1}^n \log p( y_c(t) | x(t); \theta)
\end{equation}
such that $\theta$ is the parameter-set of the model and the probability defined in Eq. (\ref{eq:image_decision}). Our infrastructure and parameter setting include TensorFlow framework using Adam optimizer with learning rate $10^{-4}$, dropout, $l_2$-regularization and a batch-size of 256 images (with 200 regions in average per image). 
To enlarge and balance the training set, we used augmentations by rotations, flips and image shifts.
\section{Experimental Results }
{\bf Dataset.} For evaluation of our method we used a large multi-center full field digital mammography (FFDM) screening dataset acquired from various devices (with approximately 3K $\times$ 1.5K image size). From this proprietary dataset we excluded images containing artifacts such as metal clips, skin markers, etc., as well as large foreign bodies (pacemakers, implants, etc.). Otherwise, the images contain a wide variation in terms of anatomical differences, pathology (including benign and malignant cases) and breast densities that corresponds to what is typically found in screening clinics.
In terms of global image BI-RADS (Breast Image Reporting and Diagnostics System), we had 350, 2364, 146 and 107 corresponding to BI-RADS 1,2,4 and 5 respectively. Global BI-RADS were set by the most severe finding in the image, according to clinical guidelines. Mammograms with global BI-RADS 3 were excluded since these intermediate BI-RADS are commonly assigned based on other modalities (\eg US) and comparison to prior mammograms \cite{excludeBIRADS3Justification2016}, which are not available in our test case. However, our data set includes BI-RADS 3 findings that were not the most severe ones in the image. We split the mammograms into the following three global labels: BI-RADS 4 \& 5 defined as malignant (M), BI-RAD 2 as benign (B) and BI-RAD 1 as normal (N).  

Our evaluation is based on 5 fold \textbf{patient-wise} cross-validation, where folds were spitted by patient IDs. We also used the INbreast (INB) public FFDM dataset \cite{moreira2012inbreast} for our evaluation. We split it into 100 positive (global BI-RADS 4,5,6) and 310 negative (global BI-RADS 1,2,3) mammograms. As this is a small data set, we conducted a random patient split on INB with 50\% for train and 50\% for test. 

{\bf Evaluation.} We compare our proposed model (Cls-Det-RS) to the previously published method of \cite{choukroun2017mammogram} (Max-region) and the DB-Baseline presenting an equivalent dual-branch approach of \cite{bilen2016weakly}. For evaluation, we present our results on two binary classification tasks, frequently used in MG, by grouping two ``nearby" classes. To this end, we used $p(y_M=1|x)$ scoring for M vs. B $\cup$ N (M vs. BN) and $max\{p(y_M=1|x),p(y_B=1|x)\}$ scoring for M $\cup$ B vs. N (MB vs. N). 

In addition to AUROC we also report two more practical measures as used in \cite{DreamChallenge}. The partial-AUC ratio (pAUCR), associated with the ratio of the area under the ROC curve in a high sensitivity range ([0.8,1]), representing the AUROC at a more relevant domain for clinicians.  Also the specificity extracted from the ROC curve at sensitivity 0.85 representing an average operation point (OP) for expert radiologists, as reported in \cite{BCSC2017}.
Table \ref{table:Results} presents the binary classification performance in two classification scenarios of M vs. BN and MB vs. N. Our method outperformed both the DB-Baseline and Max-region \cite{choukroun2017mammogram} in all measures and both classification scenarios. There was an improvement of 10-17\% on pAUCR and 8-11\% on specificity at sensitivity 0.85 ($p<0.001$). In the case of MB vs. N we obtain 6-13\% higher pAUCR and an 8-14\% higher specificity at 0.85 sensitivity ($p<0.001$). In terms of AUROC, the results showed an improvement of 2-4\% with respect to the previous Max-region and the DB-Baseline. We obtained a similar trend at breast level analysis (using maximum score between views). Train and test on the small public data set of INB yielded AUROC of 0.73. This is comparable to reported results in the literature, and close to the fully-supervised results on single view in \cite{Neeraj17} with AUROC of 0.74. Our region based error analysis indicates that the majority of our errors are between $B$ and $M$ class as it is also the case for breast radiologists. 
\begin{table}[!ht]
\begin{center}
\resizebox{\columnwidth}{!}{%
\begin{tabular}{|l| c| c| c |}
\hline
Method &  AUROC & pAUCR &  OP \\
\hline
\multicolumn{4}{c}{\bf M vs. BN weak supervision} \\
\hline
DB-Baseline \cite{bilen2016weakly} & .709$\pm$ .020 & .251$\pm$.05 & .37 \\
Max-region \cite{choukroun2017mammogram}& .699$\pm$ .047 & .235$\pm$.10 & .36  \\
Cls-Det-RS 	& \bf{.728}$\pm$ .036 & \bf{.275$\pm$.10} & \bf{.40}  \\
\hline
\multicolumn{4}{c}{\bf MB vs. N weak supervision} \\
\hline
DB-Baseline \cite{bilen2016weakly} & .826$\pm$ .01 & .347$\pm$.03 & .51 \\
Max-region \cite{choukroun2017mammogram}& .817$\pm$ .02 & .323$\pm$.07 & .48 \\
Cls-Det-RS 	& \bf{.841}$\pm$ .02 & \bf{.367}$\pm$.05 & \bf{.55} \\
\hline
\end{tabular}
}
\end{center}
\caption{Binary classification performance. OP: Specificity @ 0.85 sensitivity.}
\label{table:Results}
\end{table}

\begin{figure}[!ht]
\centering
\includegraphics[width=\columnwidth, height=2.3 cm]{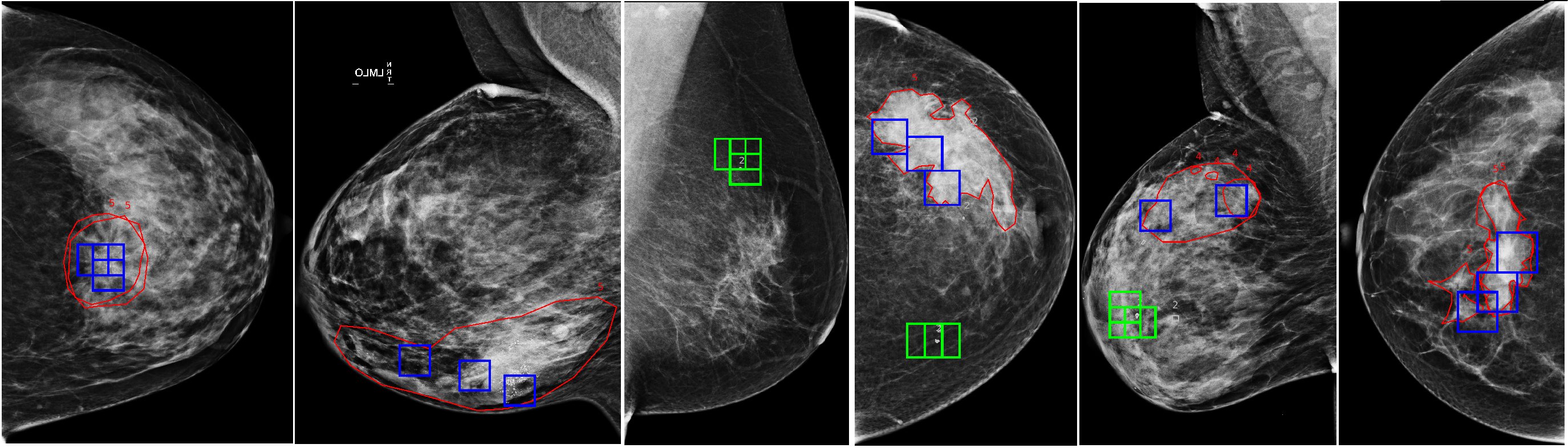}
\caption{Localization in classified images. Malignant and Benign lesions are annotated in red and gray respectively. Top 3 M (blue) and B (green) region scores are shown. Note the correlation between the radiologist annotation and the model prediction for each class.}
\label{fig:loc4fig}
\end{figure}

\begin{figure}[!ht]
\centering
\centering
\includegraphics[width=7 cm, height = 5 cm]{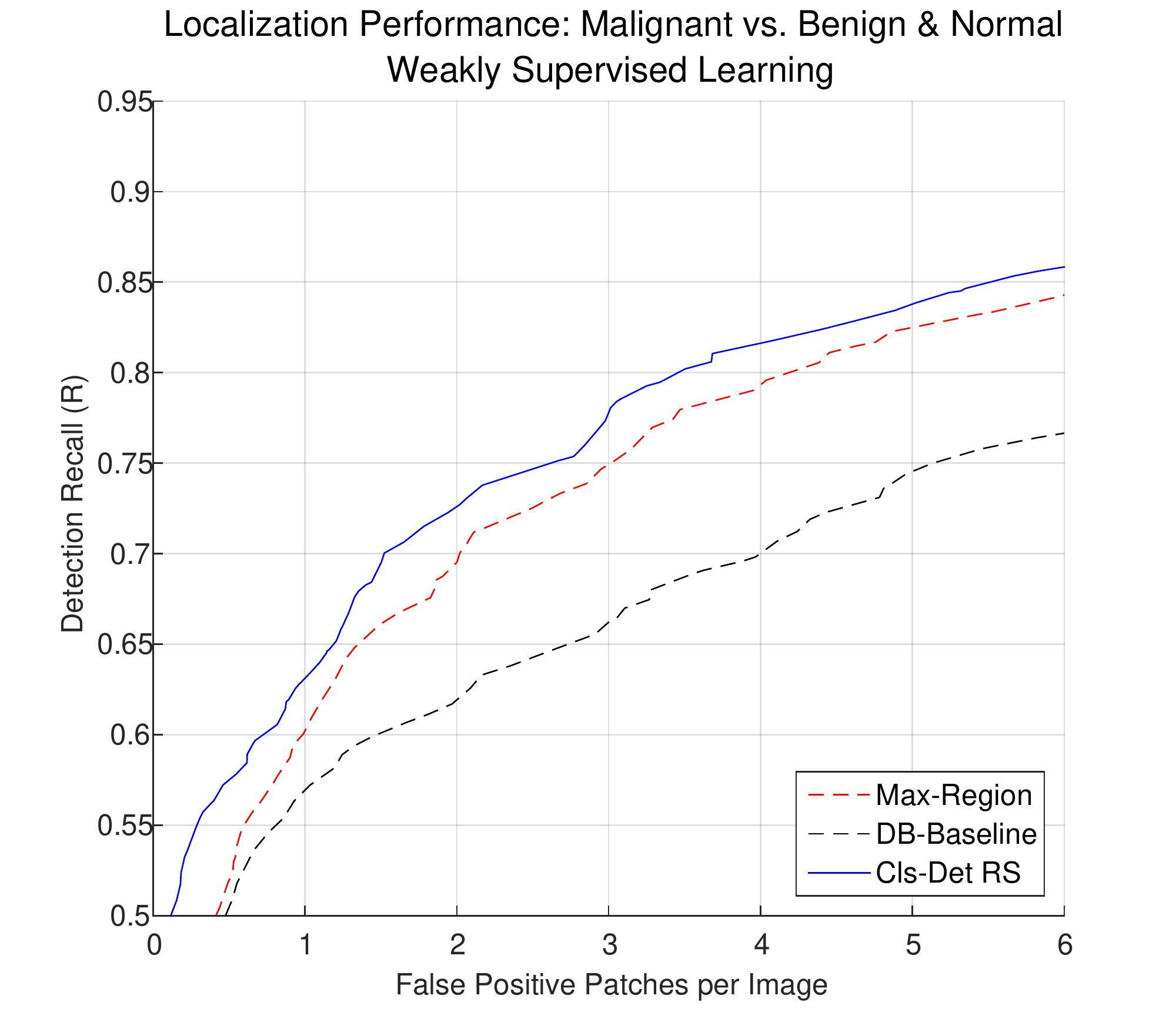}
\\ \vspace{0.25 cm}

\includegraphics[width=7 cm, height=5 cm]{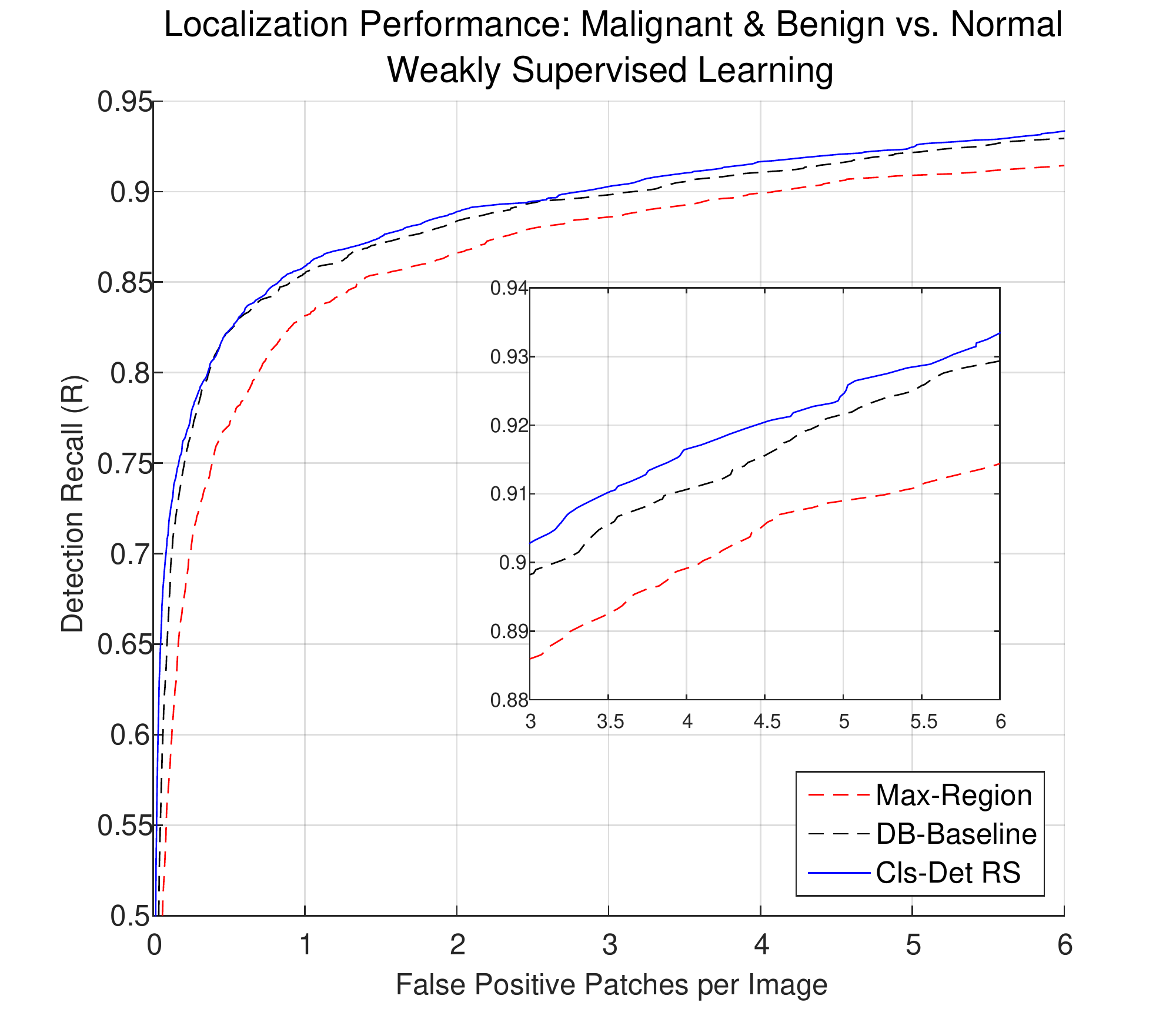}
\caption{Localization FROC at 0.85 sensitivity in classification. Note the superior performance (higher curve) of the proposed Cls-Det-RS method, particularly in M vs. BN.}
\label{fig:locRFFig}
\end{figure}

Our model further allows localization of abnormalities by highlighting the regions with top scores. We define this score as joint probability for detection (with $h_c \equiv 1$) and classification for each class:
\begin{equation}
d^c(x_i) = p_{cls}(c|x_i) p^c_{det}(i|x) \quad i \in \{1,...,m\}
\end{equation}
Figure \ref{fig:loc4fig} shows several examples with localization in the test set, overlaid with radiologist annotations (used only for validation).
As observed, the method is capable of separately highlighting benign and malignant lesions  without having an instance level annotation.
We further compare our method performance in localization to Max-region and DB-Baseline. Due to large scale factor in lesion size we use the accuracy measure Intersection over Minimum (IoM) in \cite{AD_ISBI17}:
\begin{equation}
IoM(i,c) = \frac{|r_i \cap c|}{\min \{|r_i|,|c|\}},
\end{equation}
with $c$ as the delineated domain and $r_i$ as the extracted region. We define correct localization as $IoM \geq 0.5$. This measure allows explanation of an outcome when a specified region contains a true type of lesion or vice-versa.  Note that the region size is relatively small and fixed (see Fig. \ref{fig:loc4fig}). We present the FROC localization accuracy for class $c \in \{M,B\}$ using $d^c(x_i) \geq Threshold$. The recall rate $R$ in the FROC is the fraction of images in the True-Positive set with at least one correct localization. The FROC curves in  Fig. \ref{fig:locRFFig} show the superior performance of the proposed method in localization. Note that for M vs. BN improvement is more emphasized.  This scenario sets a higher challenge since confusion between detection of M \& B  are penalized in FROC score.

To conclude, in this work, we propose a weakly supervised method for  classification and detection of abnormalities in mammograms. We consider the classification of prevalent normal cases in MG as a separate class, lacking any abnormality. Our future work includes the aggregation of extra views with bilateral and prior images, into the weakly supervised setting. These additional features together with our inexpensively labeled data and explanatory model, will close the gap towards a practical tool for radiologists. 
\newpage
\bibliographystyle{IEEEbib}
\bibliography{ref}

\end{document}